\documentclass[10pt,twocolumn,letterpaper]{article}

\usepackage{wacv}
\usepackage{times}
\usepackage{epsfig}
\usepackage{amsmath}
\usepackage{amssymb}
\usepackage{array}
\usepackage{multirow}
\usepackage{subfigure}
\usepackage{makecell}
\usepackage{float}
\usepackage{graphicx}
\usepackage{caption}
\usepackage{subcaption}
\usepackage{hyperref}
\hypersetup{
    colorlinks=true,
    linkcolor=blue,
    filecolor=magenta,      
    urlcolor=cyan,
}

\wacvfinalcopy 

\def\wacvPaperID{1384} 

\ifwacvfinal\fi
\begin{document}

\title{Re-evaluation of Face Anti-spoofing Algorithm in Post COVID-19 Era Using Mask Based Occlusion Attack}

\author{Vaibhav Sundharam \\
Virginia Polytechnic Institute and State University\\
{\tt\small vaibhavsundharam@vt.edu}
\and
Abhijit Sarkar \\
Virginia Tech Transportation Institute\\
{\tt\small ASarkar1@vt.edu}
\and
A. Lynn Abbott \\
Virginia Polytechnic Institute and State University\\
{\tt\small abbott@vt.edu}
}

\maketitle
\ifwacvfinal\thispagestyle{empty}\fi

\begin{abstract}
Face anti-spoofing algorithms play a pivotal role in the robust deployment of face recognition systems against presentation attacks. Conventionally, full facial images are required by such systems to correctly authenticate individuals, but the widespread requirement of masks due to the current COVID-19 pandemic has introduced new challenges for these biometric authentication systems. Hence, in this work, we investigate the performance of presentation attack detection (PAD) algorithms under synthetic facial occlusions using masks and glasses. We have used five variants of masks to cover the lower part of the face with varying coverage areas (low-coverage, medium-coverage, high-coverage, round coverage), and 3D cues. We have also used different variants of glasses that cover the upper part of the face.  We systematically tested the performance of four PAD algorithms under these occlusion attacks using a benchmark dataset. We have specifically looked at four different baseline PAD algorithms that focus on, texture, image quality, frame difference/motion, and abstract features through a convolutional neural network (CNN). Additionally we have introduced a new hybrid model that uses CNN and local binary pattern textures. Our experiment shows that adding the occlusions significantly degrades the performance of all of the PAD algorithms. Our results show the vulnerability of face anti-spoofing algorithms with occlusions, which could be in the usage of the algorithms in the coming era \footnote{Note: This work was done in 2020. Hence the work reflects work until that point.}.


\end{abstract}

\section{Introduction}

Face recognition systems are widely used today in mobile devices, computers, and home and office security systems for biometric identification and authentication. With growing popularity, especially due to their ease of use, different enterprises are working towards the implementation of such systems in diverse applications. These include banking, smart lock, surveillance, and access control. Although face recognition systems have reached high levels of accuracy in recognizing individuals~\cite{1}, many of these algorithms are susceptible to presentation attacks. By simply presenting a printed photograph or a video clip of a bonafide individual, an imposter could \textit{fool} the face recognition algorithm and gain unauthorized access. Some of the most common presentation attacks are print or photo attacks, video or replay attacks, and 3D mask attacks ~\cite{2,3,4}. In a print attack, a photograph of a bonafide individual is presented to the input device (e.g. webcam, front camera of mobile devices). In the replay attack, a video clip of a bonafide person is presented . For a 3D attack, a subject-specific 3D face mask is presented to the input device.\par

\begin{figure}
\centering
    \begin{subfigure}
    \centering
    \includegraphics[width=1.0\linewidth]{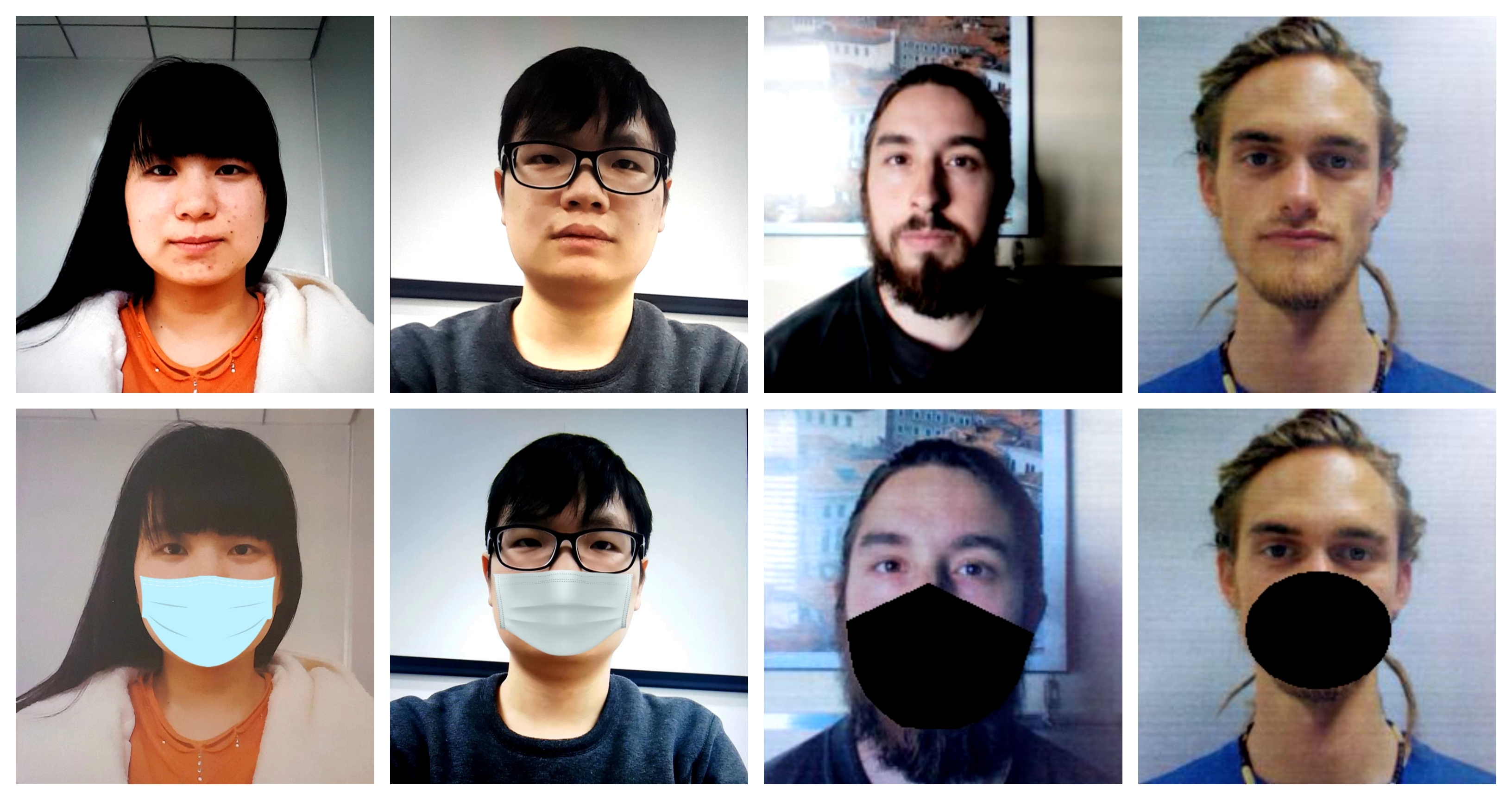}
    \label{fig:reply}
    \end{subfigure}
    \caption{Example of spoof attacks with mask occlusions. First row depicts \textit{presentation} attack images and second row depicts  occlusion attacks by different kinds of masks. }
    \label{fig:database examples}
\end{figure}

All these Presentation attack instrument (PAI)
that are used in presentation attacks~\cite{Kolberg2020}, generate artifacts. The goal of most of the legacy face anti-spoofing algorithms is to identify these artifacts PAIs. This may include, change in texture pattern between the real image and the attacked images, quality degradation during sample recapture~\cite{2,3,5}, and lack of 3D cues. Therefore, the standard practice to develop such anti-spoofing algorithms was to first extract features that can capture such discrepancies and then run a binary classifier to train a model.  Over the years, Local Binary Patterns (LBP) was widely used ~\cite{6,2,22} to perform texture-based anomaly detection. Other notable PAD algorithms leverage color~\cite{7}, motion~\cite{8}, or subject’s liveness~\cite{9} to generate distinctive features for PAD algorithms.
In recent years, due to the popularity and success of convolution neural networks (CNNs), several PAD algorithms based on CNNs have been proposed~\cite{10,11,12,13}. Apart from raw RGB images, some researchers have proposed auxiliary supervision, such has depth maps~\cite{14}, rPPG signals~\cite{5}, pixel-wise binary supervision~\cite{15}, or a combination of these, to boost the performance of the CNN based PAD algorithms and make them generalize well.
The biggest advantage of a CNN based PAD algorithm is that it does not require handcrafted features as compared to previous PAD algorithms. Instead, features are learned through training a task specific end-to-end CNN. \par
Now, the decision of any algorithms depends on specific image properties that can be retrieved from an image. Occlusion often plays pivotal role in obscuring some of the key image features. 
Zeiler et al \cite{zeiler2014visualizing} has shown that occlusion to a particular part of an object can effect its final classification decision of a network. Even a single pixel adversarial attack can \textit{fool} a CNN \cite{su2019one}. Prior research work on occlusion in biometric shows that blocking areas in the face degrades performance of PAD algorithms \cite{22}. A mask or a sun-glass occludes part of the face, hence some information is lost. How that loss of information affects the decision boundary of a classifier is unknown.

\subsection{Face biometric algorithms in post pandemic era}
The coronavirus disease (COVID-19) caused by the SARS-CoV-2 virus~\cite{16} has put the world in a state of emergency. Governments across the globe have mandated the use of  protective face masks in public places. As most of the benchmark dataset for face recognition and face anti-spoofing algorithms do not use  occluded face databases in their training, questions are raised in their viability in post pandemic era. As a result, any face biometric algorithms, including face recognition and face anti-spoofing algorithms, developed using those benchmark data are under a scanner. 
Recently, the work of Ngan \etal~\cite{41} has demonstrated that the addition of masks to face images significantly affect the performance of pre-COVID \textit{face recognition} algorithms. They observed a significant increase in the False Rejection Rate (FRR) by 20\%-50\% on state of the art face recognition algorithms which are tested with mask occlusions. Their experiments shows increase in False Acceptance Rate (FAR) in most of the face recognition algorithms. Their work clearly shows the vulnerability of pre-COVID face recognition algorithms. \par

Due to similar research practice in the face anti-spoofing or alternatively PAD algorithms in the biometric literature it becomes imperative to perform a similar set of experiments on these algorithms and re-evaluate their performance in the wake of this pandemic. Motivated by this, we analyze the performance of PAD algorithms in the presence of occlusions, that covers different part of the face. We have used the mask occlusion, effecting the lower part of the face, with four different coverage areas (low-coverage, medium coverage, large coverage, round shaped coverage), and different level of 3D cues (different in shades, patterns,and albedo). We have also used different glasses that partially occludes the upper part of the face including eyes. Examples of spoof attacks with occlusions are shown in Figure~\ref{fig:database examples}. The first row depicts the images of presentation attacks whereas the second row depicts occlusions applied to the presentation attack images.  

We have specifically focused on four different PAD algorithms that were developed based on texture, frame difference, image quality, and CNN features. We used images and videos from two benchmark datasets, Replay-Attack~\cite{2} and OULU-NPU~\cite{4}, and applied synthetic occlusion attacks. 
In the process, we investigate the following: \par
\begin{itemize}
    \item How the PAD algorithms perform in the presence of different occlusion attack?
    \item Does the amount of the face occlusion have an impact on the performance of the PAD algorithm?
    \item Is there any performance difference if masks having 3D cues are used instead of a flat 2D mask occlusion?  
    \item How PAD algorithms based on hand crafted feature such as texture, frame difference, image quality performs in comparison to CNN based PAD model?
\end{itemize}

The paper is organized as follows. Section (\hyperref[sec:Literature-Survey]{2}) presents the literature survey of various PAD algorithms used in face anti-spoofing along with popular techniques used to develop such algorithms. In section (\hyperref[sec:Experiments]{3}), we go over the experimental set-up, detailing how the occlusion attacks were synthesized, here we introduced the hybrid methods that we developed in order to bridge the gap between the traditional and CNN methods.  In section (\hyperref[sec:Discussion]{4}), where we discuss the results of the occlusion attack on the baseline PAD systems. Finally, in section (\hyperref[sec:Conclusion]{5}), we present the conclusion as well as possible future work. To the best of our knowledge, this is the first work that evaluates pre-COVID PAD algorithms with mask occlusion attack.
\section{Literature Survey}
\label{sec:Literature-Survey}
In this section, we first discuss different PAD methods commonly used, followed by a brief overview of occlusion attack, including mask-based attacks in biometric literature.  Broadly speaking, most of the PAD algorithms can be segregated into two categories, i.e. appearance-based, and temporal-based methods.

\subsection{Appearance-based methods}
Print and video attacks tend to leave some artifacts on the image that may trigger a PAD algorithm. Much of the prior research in this field exploits this fact for spoof detection. The use of handcrafted features such as Local Binary Pattern (LBP)~\cite{2,6}, LBP-Lacunarity~\cite{22},  color space~\cite{7}, image distortion~\cite{19}, and image quality~\cite{21} along with traditional classification algorithms like Support Vector Machines (SVM) and Linear Discriminant Analysis (LDA) are extensively studied and documented. With the success of neural networks, especially in computer vision problems, there have been a lot of efforts in developing CNN based feature extractors~\cite{11,13,24,25}. Moreover, to improve the performance of such algorithms, researchers have started introducing auxiliary supervision methods such as depth maps, rPPG signals, binary supervision, infrared, etc. with CNNs. For example, in~\cite{14}, a two-stream CNN architecture is proposed, where the first CNN stream extracts local features from the image, while the other is used to estimate depth maps. A fusion of CNN and LBP is proposed in~\cite{27}, where convolution feature maps are used to extract color LBP patterns. The extracted features are used with SVM classifier to differentiate real and fake faces. In~\cite{15}, authors have used pre-trained DenseNet~\cite{28} architecture along with pixel-wise supervision to generate presentation attack scores for every frame of the attack video. In George \etal~\cite{13}, authors have used multiple channels such as color, depth, and infrared to design a robust PAD algorithm for detecting 2D and 3D presentation attacks.\par 

\subsection{Temporal-based methods}
Temporal methods~\cite{8,23,29,30} based on eye blinking, mouth, and lip movement were one of the earliest approaches used to detect print attacks. These methods usually depend upon a sequence of images to detect temporal changes in the sample. For example, in~\cite{31}, authors have used the blinking pattern of eyes in humans to detect liveliness. Their model is based on the fact that humans, on average, blink at least once every two-to-four seconds. Additionally, features such as optical flow~\cite{32,33} and motion magnification~\cite{34,23} are also used to perform face anti-spoofing. Though these methods are effective in detecting print attacks, their performance is hampered when the algorithm is exposed to replay attacks. With the advent of deep learning, researchers have started looking into a fusion of CNNs and temporal features. For example, in~\cite{35}, neural networks were used with optical flow maps to perform binary classification. An LSTM-CNN architecture is deployed in~\cite{36}, which leverages the temporal information for PAD detection. Additionally, in Liu \etal~\cite{5}, authors have used a CNN-RNN architecture that learns to estimate the face depth and rPPG signals for face PAD detection. 


\section{Experiment}
\label{sec:Experiments}
In the following section, we discuss in detail the databases, synthetic occlusion attacks, and baseline algorithms used in this paper. 

\subsection{Databases}
In our experiments, we have used two benchmark databases, namely Replay-Attack~\cite{2} and the OULU-NPU~\cite{4}. These databases consist of a wide variety of ground truth videos of bonafide and imposters, trying to access an authenticating system (mobile device, laptop). The datasets are recorded under various lighting conditions and background scenes. We chose the Replay-Attack database as many of the legacy face PAD algorithms have used it to evaluate their performance. OULU-NPU, on the other hand, is a more recent dataset that is gaining popularity among biometric researchers primarily due to the inclusion of real-world variations in the data samples. Most of the CNN based algorithms use this dataset to demonstrate their generalization capability.\par
The Replay-Attack database consists of 1300, 720p high-definition video clips of presentation attacks, such as print attack and video attack. There are a total of 50 subjects that try to access a laptop either through a built-in webcam directly or by displaying a photo of themselves or by presenting a short video clip of themselves to the camera. Each of the training and development set consist of 60 real-access and 300 imposter-access videos, whereas the test set consists of 80 real-access and 400 imposter-access videos. The OULU-NPU dataset is a 1080p high-definition dataset with 990 bonafide and 3960 attack videos. It includes both print and video attacks. The front camera of mobile devices such as Samsung Galaxy S6 Edge, HTC, etc. was used to capture the video clips. Additionally, it consists of four different protocols. The first three protocols capture various conditions which include unseen environmental conditions (illumination, background scenes), different attack devices, and input camera variations respectively. The fourth protocol consists of all the aforementioned environmental conditions. In this work we have used protocol 1 to demonstrate occlusion attack. 

\subsection{Occlusion attack}
\begin{figure}
\begin{center}
   \includegraphics[width=0.7\linewidth]{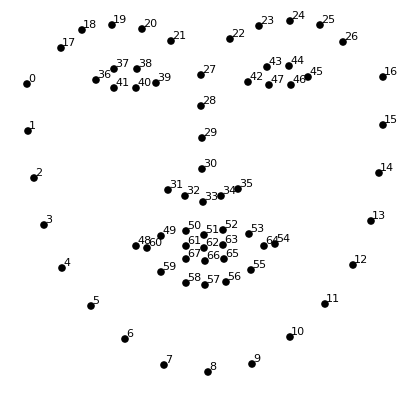}
\end{center}
   \caption{Example of the 68 facial landmarks predicted by the "dlib"~\cite{46} library.}
\label{fig:dlib}
\end{figure}

In this section, we present different types of occlusion attacks that are performed on the baseline PAD algorithms and also give a description of how these attacks were executed. Each of these attacks occludes a portion of the face in the testing data of the Replay-Attack~\cite{2} and the OULU-NPU~\cite{4} dataset by either synthesizing or adding an artifact to it. To detect the facial key points, we have used Python’s “dlib”~\cite{46} library, which consists of pre-trained models to detect faces and estimate the location of key points on the detected faces. An example of the 68 key points estimated by the “dlib” library is shown in Figure~\ref{fig:dlib}. Each of the videos in the datasets is preprocessed in the following manner. Firstly, we extract each frame from the video and feed it to the “dlib” face detection model. The model then estimates the location of the 68 landmarks. Using these landmarks as reference points, we  add synthetic occlusions to the face. 

           \begin{figure}
            \centering
            \includegraphics[width=1.0\linewidth]{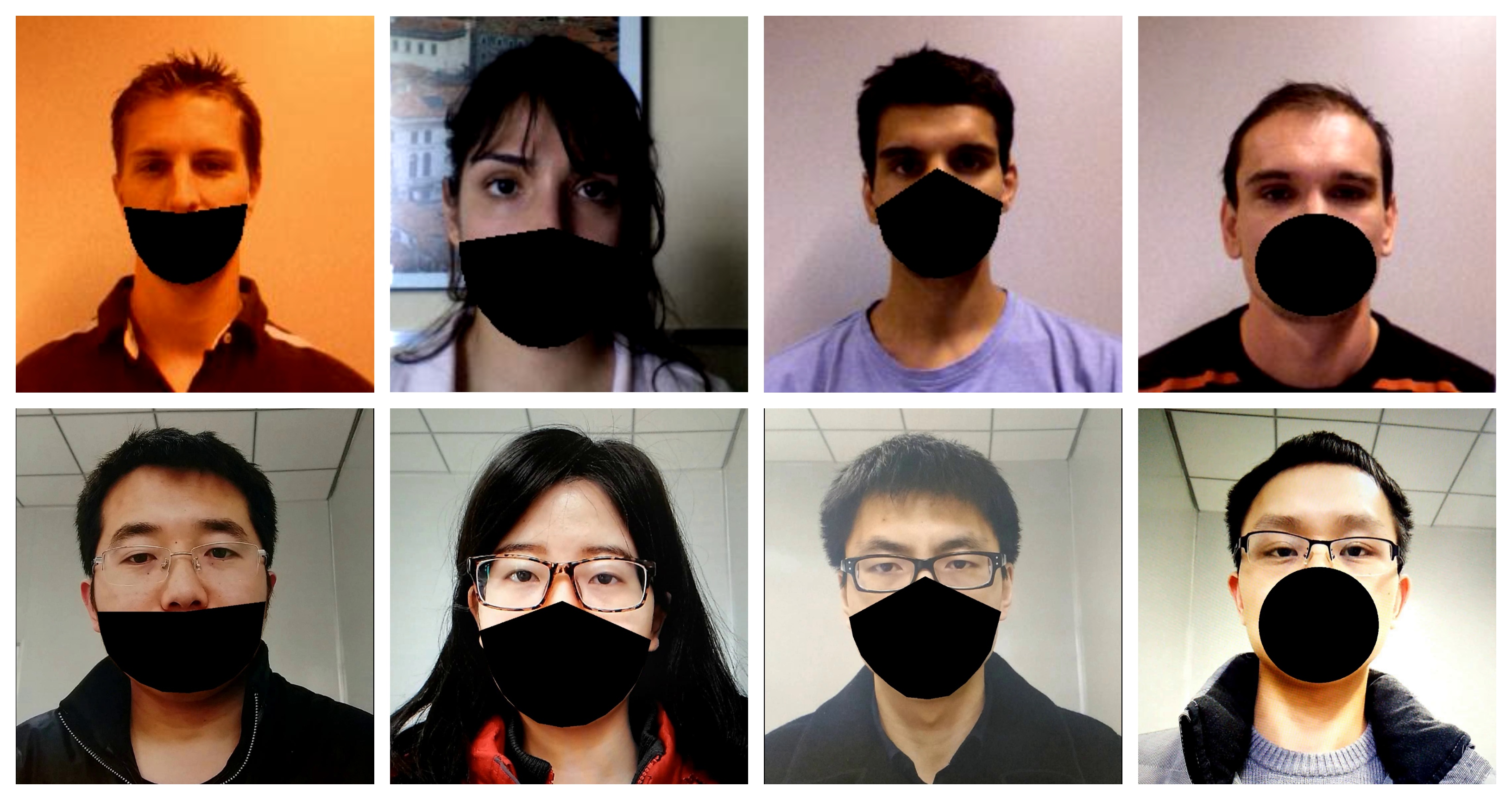}
            \caption{Figure depicting various 2D mask occlusions applied to the Replay-Attack dataset ~\cite{2} and OULU-NPU dataset ~\cite{4}. Moving from left to right, the first column depicts a 2D mask with low-coverage ($\approx 30\%$) facial coverage). The mask covers the lower part of the face, and the nose of a person is visible. The second column depicts a 2D mask with medium-coverage ($\approx 40\%-50\%$ facial coverage) with some portion of the nose occluded. The third column depicts a 2D mask with high-coverage ($\approx 50\%-70\%$ facial coverage). One can see that a complete nose is blocked in this attack. Finally, the fourth column represents a 2D mask with round coverage ($\approx 30\%-40\%$ facial coverage).}
            \label{fig:mask_2d}
            \end{figure}

    \begin{figure}
            \centering
            \includegraphics[width=1.0\linewidth]{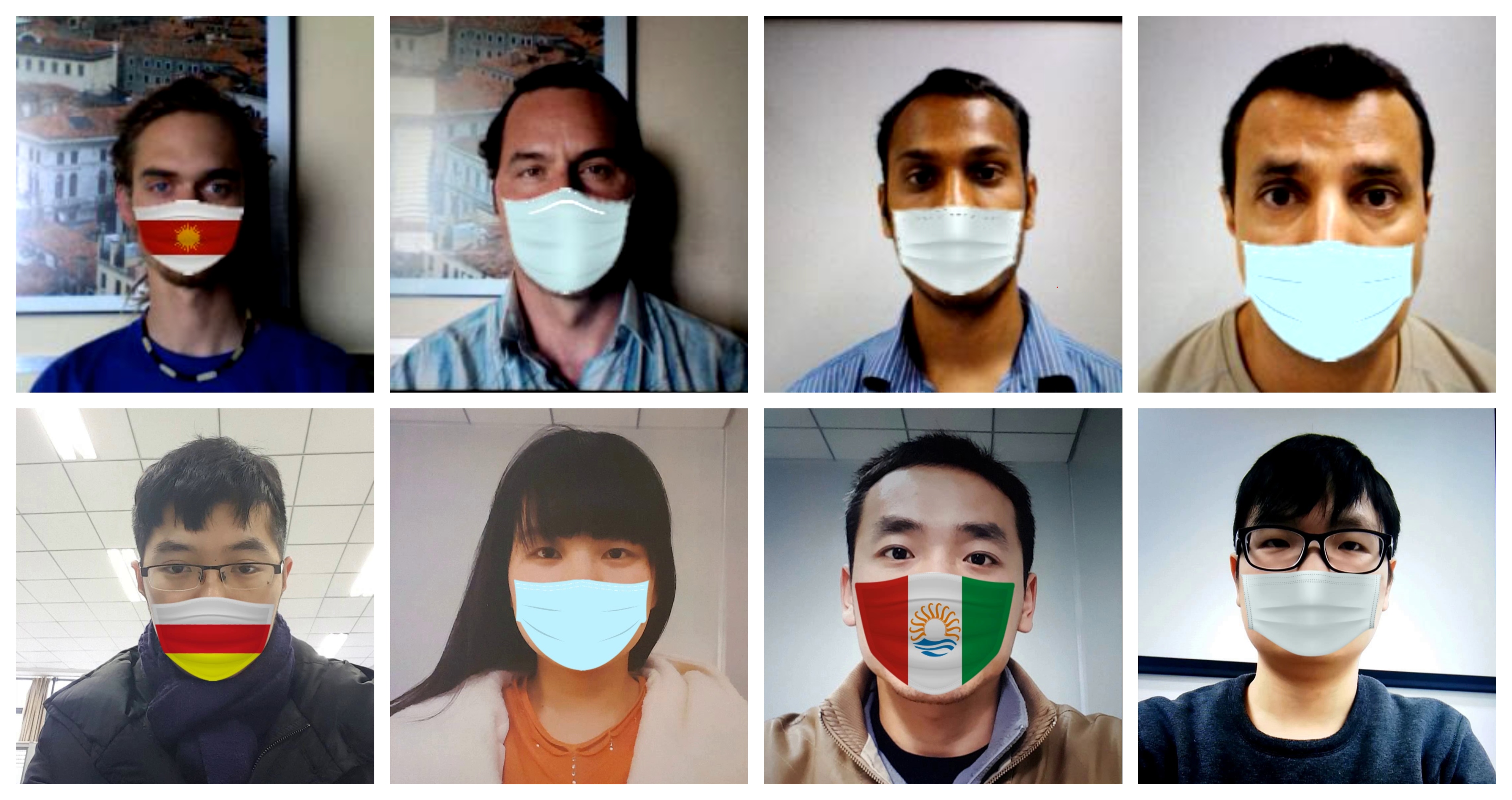}
            \caption{Figure depicting examples of mask occlusions with 3D cues on Replay Attack~\cite{2} dataset and OULU-NPU ~\cite{4} dataset.}
            \label{fig:mask_3d}
            \end{figure}


    \subsubsection{Mask attacks} 
    Inspired by the widespread use of medical masks due to the COVID-19 pandemic, we present five different types of mask attacks namely 3D mask attack, low-coverage mask attack, medium-coverage mask attack, high-coverage mask attack, and round mask attack. Figure~\ref{fig:mask_2d} and Figure~\ref{fig:mask_3d} present examples of 2D and 3D mask attacks respectively. The 2D masks add a single color artifact on the image. Also, three different face coverages (low, medium, high) are introduced in 2D mask attacks, keeping in mind the various preferences individuals have while wearing masks. The 3D masks have some 3D cues, such as texture, color-patterns, associated with it. In our experiments we have used 9 different types of 3D masks to perform the occlusion. We have also included round masks in our analysis that resemble K95 and KN95 masks.
            
    \subsubsection{Glasses attack}
\begin{figure}
    \centering
    \includegraphics[width=1.0\linewidth]{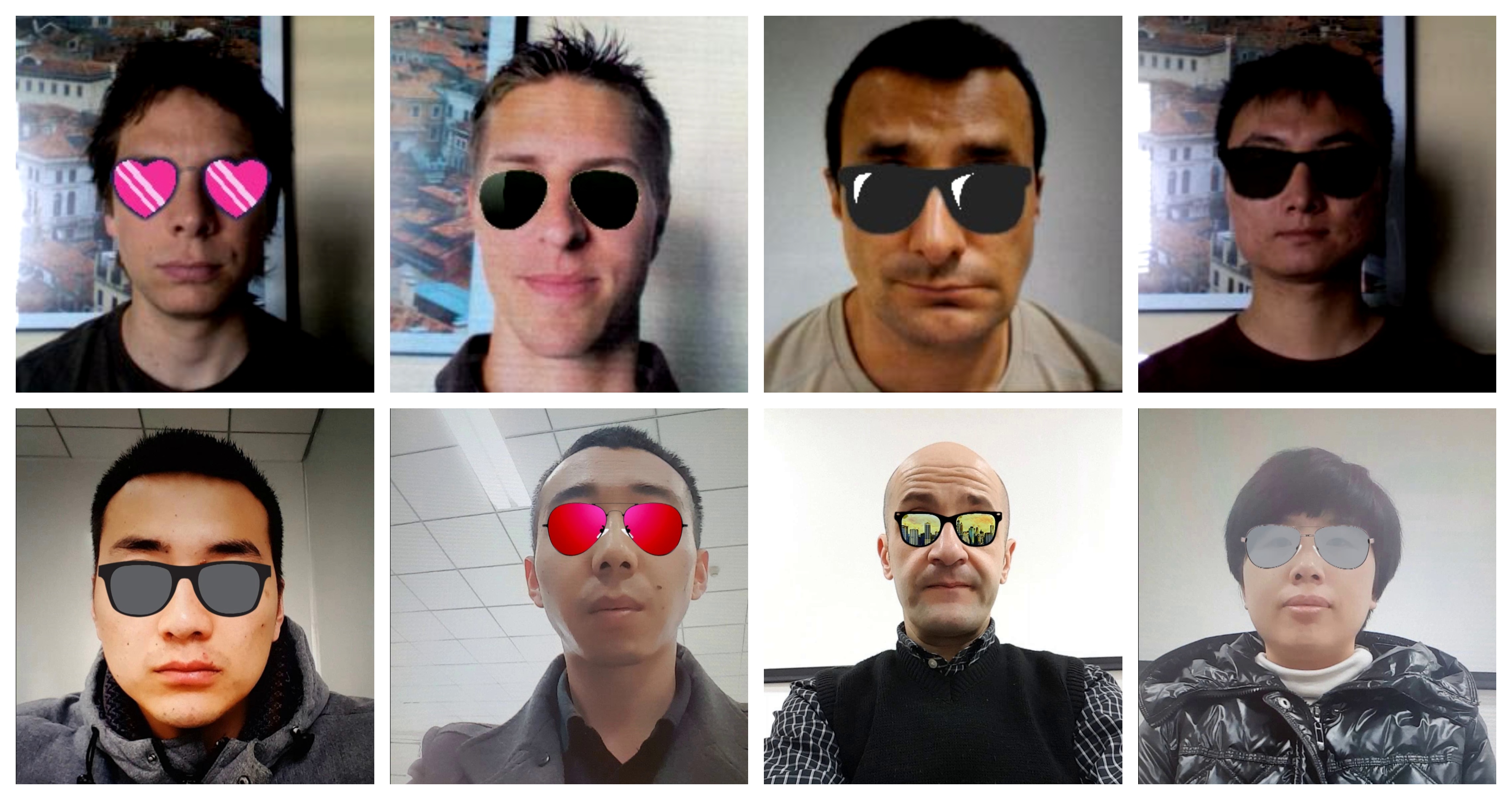}
    \label{fig:sunglasses}
    \caption{Figure depicting an example of the glasses occlusion attack on Replay Attack (a)~\cite{2} dataset and OULU-NPU (b)~\cite{4} dataset.}
    \label{fig:sunglasses_3d_oulu}
    \end{figure}

    We synthetically add glasses to the faces as shown in Figure~\ref{fig:sunglasses_3d_oulu}. The glasses have different shapes, sizes, and colors and occlude the eye region of the image. Some of these occlusions are opaque and completely block the eyes while others are translucent. The principal objective behind selecting glasses as one of the occlusion attacks was to observe the effect of such occlusion on the False Acceptance Rate of an algorithm. The motivation stems from the fact that many of the subjects in the training and development set of both the databases are wearing reading glasses. Since these algorithms are trained on these datasets (thereby retaining some information about glasses), we wanted to evaluate whether the addition of such occlusion results in the algorithm accepting an imposter as a bonafide individual. For evaluating the baseline models, we have used 12 different types of glasses.
    
\subsection{Baseline systems}
Four reproducible and publicly available face anti-spoofing algorithms were attacked using various occlusions. Each of the chosen algorithms demonstrate a different approach to detect imposter images. In addition to that we introduced a new hybrid model that interjects traditional LBP features into a CNN model in the anticipation to learn additional textural features. \par
\subsubsection{Texture based algorithm}
Bonafide and spoofed images have some subtle texture differences that may go unnoticed even to the human eyes. But, with proper translation to an appropriate feature space, these subtle differences can be realized. LBP is a popular technique used to capture texture information in an image by transforming the feature space. Hence, the first algorithm we experiment on uses texture features based on Local Binary Patterns (LBP-SVM)~\cite{2,43} at its core to perform the classification tasks. Each frame extracted from the video is preprocessed by cropping the face from each of the frames. Faces with the size less than a set threshold are discarded. After pre-processing, an LBP histogram is extracted from each frame. The SVM algorithm is used with RBF kernel to classify each frame in the video as bonafide or attack~\cite{43}.\par
\subsubsection{Iamge quality measures}
Image Quality Measure (IQM), introduced by~\cite{6671991}, is known to improve the performance of biometric recognition systems by adding liveness assessment to the frameworks. 
Hence, the second baseline algorithm presented in this paper uses the Image Quality Measure (QM-SVM)~\cite{45,43,44} as a feature vector to perform the classification task. The videos are pre-processed in a similar fashion as done in the first algorithm. The IQM features extracted from the videos are fed to an SVM classifier. \par
\subsubsection{Motion features}
Optical flow estimators are typically used for motion-based face PAD algorithms but a simpler method for stationary recognition systems, as pointed out in~\cite{8}, is to find correlation between the movement of the face and the scene background. Therefore, the third baseline algorithm used in this paper is based on frame difference (Motion+SVM)~\cite{8,43}. In the pre-processing stage, frame differences (FD) are computed for both facial and background regions, followed by a feature extraction stage, in which five features are extracted from the frame difference signal. These extracted features are used with the SVM algorithm for classification.\par
\subsubsection{CNN features}
The ability of neural networks to synthesize features is a huge advantage over other face PAD algorithms that use hand-crafted features.
To dive deeper into effects of occlusion attack on CNN based face PAD algorithims, we chose the fourth algorithm based on DenseNet~\cite{28} architecture and pixel-wise auxiliary supervision~\cite{15,47} (Deep-PiXBiS). This algorithm uses frame-level information to classify bonafide and imposter images, which results in faster processing and quicker authentication. Also, this algorithm shows good generalization in cross dataset experiments. Face detection and localization are performed on each of the input images, followed by face alignment and resizing. The DenseNet model pre-trained on ImageNet dataset~\cite{deng2009imagenet} is used. The decision threshold for carrying out the classification is based on the equal error rate criterion.

\subsubsection{Hybrid CNN Network}
We also present our hybrid algorithm based on convolution neural networks that leverage the RGB and LBP information to perform presentation attack detection. We have used 3 x 3 uniform LPB to compute the local binary patterns. This method leverages both handcrafted as well as CNN generated features to carry out the classification task. RGB images, along with their corresponding LBP maps, are fed to the neural network during training. The model is based on Resnet architecture, pre-trained on the ImageNet dataset. Un-occluded images are used to train the model whereas, images occluded with 3D masks and glasses are used while testing to see the efficacy of our model towards the occlusion attacks. The model was trained for 20 epochs with a learning rate of 0.001 and a batch size of 256.

\subsection{Performance Metric}
For the OULU-NPU~\cite{4} dataset, we adopt the newly standardized ISO/IEC 30107-3 performance metric for the evaluation of the results, first of which is the APCER, or Attack Presentation Classification Error Rate, that defines the fraction of attack presentations that were falsely classified as bonafide presentation. Secondly, we have used BPCER, or Bonafide Presentation Classification Error Rate, that gives the error in classifying bonafide images. Finally, ACER, or Average Classification Error Rate, is used which is the average of APCER and BPCER~\cite{37,15}.

\begin{equation}
    ACER = \frac{APCER+BPCER}{2}
\end{equation}

For the Replay Attack dataset~\cite{2}, we present the results using commonly adopted performance metric such as FRR or False Rejection Rate, FAR or False Rejection Rate, and HTER or Half Total Error Rate.  HTER is defined as the mean of FAR and FRR. Additionally, EER, or Equal Error Rate computed from the development set, is used as a decision threshold.

\begin{equation}
    HTER = \frac{FRR+FAR}{2}
\end{equation}
\section{Results}
\label{sec:Discussion}

\begin{table*}
\small
\begin{center}
\begin{tabular}{ |c|c|c|c|c|c|c|c|c| } 
\hline
\textbf{Baseline} & \textbf{Metric} & \textbf{\makecell{No occlusion}} & \textbf{\makecell{Low\\coverage}} & \textbf{\makecell{Medium\\coverage}} & \textbf{\makecell{High\\coverage}}  & \textbf{\makecell{Round\\coverage}}  & \textbf{3D mask} & \textbf{Glasses} \\
\hline
\multirow{3}{7em}{\makecell{LBP+SVM~\cite{2}}} & FAR & \makecell{9.4\%} & \makecell{0.9\%} & \makecell{0.9\%} & \makecell{0.9\%} & \makecell{0.9\%} & \makecell{26.9\%} & \makecell{27.5\%} \\ \cline{2-9}
& FRR & \makecell{21.8\%} & \makecell{98.7\%} & \makecell{98.7\%}  & \makecell{98.7\%} & \makecell{98.7\%} & \makecell{54.6\%} & \makecell{30.3\%} \\ \cline{2-9}
& HTER & 15.6\% &  49.8\% & 49.8\% & 49.8\% & 49.8\% & 40.8\% & 28.9\% \\ 
\hline

\multirow{3}{7em}{\makecell{QM+SVM~\cite{44}}} & FAR & \makecell{7.8\%} & \makecell{0.6\%} & \makecell{3.2\%} & \makecell{8.9\%} & \makecell{3.4\%} & \makecell{13.2\%} & \makecell{10.9\%}\\ \cline{2-9}
& FRR & \makecell{1.3\%} & \makecell{68.3\%} & \makecell{84.0\%}  & \makecell{83.8\%} & \makecell{84.3\%} & \makecell{85.4\%} & \makecell{44.4\%} \\ \cline{2-9}
& HTER & 4.6\% &  34.4\% & 43.6\% & 46.3\% & 43.9\% & 49.3\% & 27.6\% \\ 
\hline

\multirow{3}{7em}{\makecell{FD+SVM~\cite{8}}} & FAR & \makecell{13.9\%} & \makecell{47.9\%} & \makecell{45.8\%} & \makecell{45.7\%} & \makecell{45.9\%} & \makecell{40.3\%} & \makecell{57.3\%} \\ \cline{2-9}
& FRR & \makecell{12.5\%} & \makecell{7.7\%} & \makecell{9.0\%}  & \makecell{10.0\%} & \makecell{9.3\%} & \makecell{48.2\%} & \makecell{4.3\%} \\ \cline{2-9}
& HTER & 13.2\% &  27.8\% & 27.4\% & 27.8\% & 27.6\% & 44.3\% & 30.8\% \\ 
\hline

\multirow{3}{8em}{\makecell{Hybrid CNN(Ours)}} & FAR & \makecell{3.8\%} & \makecell{0.24\%} & \makecell{0.24\%} & \makecell{0.24\%} & \makecell{0.24\%} & \makecell{0.23\%} & \makecell{1.67\%} \\ \cline{2-9}
& FRR & \makecell{0\%} & \makecell{94.2\%} & \makecell{94.0\%}  & \makecell{93.8\%} & \makecell{92.2\%} & \makecell{93.0\%} & \makecell{17.05\%} \\ \cline{2-9}
& HTER & 1.9\% &  47.22\% & 47.12\% & 47.06\% & 46.22\% & 46.6\% & 9.36\% \\ 
\hline
\end{tabular}
\end{center}
\caption{The results of mask occlusion and glasses occlusion attack on based baseline algorithms, tested on Replay-Attack~\cite{2} }
\label{table:mask_attack}
\end{table*}

Table~\ref{table:mask_attack} depicts the performance metric values of LBP+SVM, QM-SMV and, Motion+SVM face PAD algorithms when attacked with mask occlusions. LBP+SVM, QM+SVM, and Motion+SVM are trained on the Replay-Attack dataset with unoccluded 300 attack videos and 60 real videos. For the SVM based algorithms, testing is done with occluded images on the “grandtest” protocol as specified in~\cite{2,15,43,44}. It should be noted that out of all testing video clips in the Replay-Attack dataset, the "dilb" library was unable to detect and predict landmarks on 20 videos. Hence, these 20 videos are fed to the model without adding any occlusions. For the Deep-PiXBiS model, we have used the pre-trained weights of the model trained on Protocol 1 of the OULU-NPU dataset to test the performance of the architecture with occluded test images. Table~\ref{table:sunglasses_attack} summarizes the performance of the CNN based algorithm (Deep-PiXBiS) when attacked by various occlusions. 

\begin{table*}
\small
\begin{center}
\begin{tabular}{ |c|c|c|c|c|c|c|c|c| } 
\hline
\textbf{Baseline} & \textbf{Metric} & \textbf{\makecell{No\\occlusion}} & \textbf{\makecell{Low\\coverage}} & \textbf{\makecell{Medium\\coverage}} & \textbf{\makecell{High\\coverage}}  & \textbf{\makecell{Round\\coverage}}  & \textbf{3D mask} & \textbf{Glasses} \\
\hline
\multirow{3}{9em}{\makecell{Deep-PiXBiS~\cite{15,47}}} & APCER & \makecell{0.83\%} & \makecell{0.42\%} & \makecell{2.5\%} & \makecell{2.5\%} & \makecell{2.5\%} & \makecell{2.5\%} & \makecell{16.67\%}\\ \cline{2-9}
& BPCER & \makecell{0.0\%} & \makecell{7.5\%} & \makecell{10.0\%}  & \makecell{11.0\%} & \makecell{7.5\%} & \makecell{39.17\%} & \makecell{12.5\%} \\ \cline{2-9}
& ACER & 0.42\% &  3.96\% & 6.25\% & 6.75\% & 5\% & 20.83\% & 14.58\% \\ 
\hline
\end{tabular}
\end{center}
\caption{The results of mask occlusion and glasses occlusion attack on CNN based baseline algorithms tested on OULU-NPU~\cite{4}}
\label{table:sunglasses_attack}
\end{table*}

\subsection{Performance of LBP+SVM model}
Testing the model without adding any occlusion gives performance with a FAR of 9.4\%, FRR of 21.8\%, and an HTER of 15.6\%.  As we add 2D mask occlusions (low-coverage, medium-coverage, high-coverage, round-coverage) to the test images, the performance of the algorithm severely deteriorates. Also, it can be seen that the model’s performance is the same across all four aforementioned occlusions i.e, increasing the percentage of the area occluded does not effect the performance when it comes to 2D occlusions. But, in all cases, the FRR increases from 21.8\% to 98.7\%. The FAR, on the other hand drops from 9.4\% to 0.9\%. This in turn shows that with 2D masks, the algorithm rejects even a real person.  Introduction of the 3D cues significantly increases both FAR and FRR compared to when no occlusion is applied to the test images. For the 3D mask attack, we see an increase of 17.5\% in the FAR whereas, for glasses, we see an increase of 18.1\% in the FAR. From the observation, it can be concluded that the additional information imparted by the textures in a 3D mask attack is able to spoof the LBP+SVM algorithms into accepting more imposter images as genuine. A similar argument can be drawn for the increase in FAR that we see after the addition of glasses.

\subsection{Performance of QM+SVM model}
Testing this PAD algorithm without adding any occlusion to the test images results in the baseline HTER of 4.6\% with FAR and FRR as 7.8\% and 1.3\% respectively. The percentage of the image occluded by the 2D mask seems to correlate with the model's overall performance. As the percentage of area occluded by the mask increases from low-coverage to high-coverage the HTER of the system increases from 34.4\% to 46.3\%, with the HTER of round-coverage lying in between. Similar to the LBP+SVM model, the highest increase in the FRR and FAR is seen when masks with 3D cues are applied to the test images as compared to when no occlusion is applied. The addition of glasses also deteriorates the performance of the algorithm with the increase in the HTER from 4.6\% to 27.6\%. 

\subsection{Performance of Frame Difference + SVM}
For this algorithm, testing with un-occluded images gives 13.9\% as the FAR and 12.5\% as the FRR. With occlusion, we see that there is a substantial increase in the False Acceptance Rate across all types of mask occlusions. This observation is different as compared to the algorithms discussed above where adding occlusions deteriorates FRR more than the FAR. Adding glasses results in an increase in the FAR from 13.9\% to 57.3\%. Additionally, the introduction of 3D masks to the test data has the most significant impact on the False Rejection Rate of the algorithm. 

\subsection{Performance of Deep-PiXBiS}
For this CNN based algorithm, the APCER and BPCER on the test set, without adding any occlusion, is 0.83\% and 0\% respectively~\cite{15,47}. The overall performance of the system deteriorates as we move from low-coverage occlusion to high-coverage occlusion for the 2D mask attack, which is expected. The increase in the APCER is not significant with the maximum increase of only 1.67\%. Additionally, the APCER does not deteriorate much on the addition of 3D masks but, there is a substantial increase in the BPCER from 0\% to 39.1\%. Interestingly, the addition of glasses to the test images is increasing the APCER from 0.83\% to 16.67\%  which roughly translates to a 20 times increase in the APCER. The BPCER of the PAD system is also affected under this attack with an increase from 0\% to 12.5\%.

\subsection{Performance of Hybrid CNN Network}
On testing this model without any occlusions gives an HTER of 1.9\%. This is the best results for the replay-attack dataset reported in this paper. However, on adding mask occlusions, the performance of the system severely drops. This is attributed to the fact that the FRR increases from 0 to $\approx 93\%$. Even though adding occlusions is deteriorating the overall performance of the model, we see that there isn't a significant increase in FAR. Hence, for Replay-Attack dataset, this hybrid CNN based model proves to be more robust towards the false acceptance of imposter images.

\section{Discussion}
The results of the benchmark algorithms evidently shows that occlusions affect the performance of all the algorithms. Although, we can find some interesting patterns in the failure of such methods. First, the CNN based methods seem to be most robust amongst the algorithms. This reemphasises the importance of CNN in vision based research. Second, face coverage areas mostly effects the FRRs in texture based models (LBP+SVM and QM+SVM). With higher coverage area, FRR increases. Third, this is in contrast to the motion based models (FD+SVM) which deteriorate mostly in FAR. Fourth, FAR of CNN is affected most by the glass attack. This intuitively indicates its vulnerability to spoofing while occluding upper part of the face. Fifth, the hybrid method introduced in this paper is most robust for FAR and for the glass attack.  \par

\begin{figure}
    \begin{center}
        \includegraphics[width=1.0\linewidth]{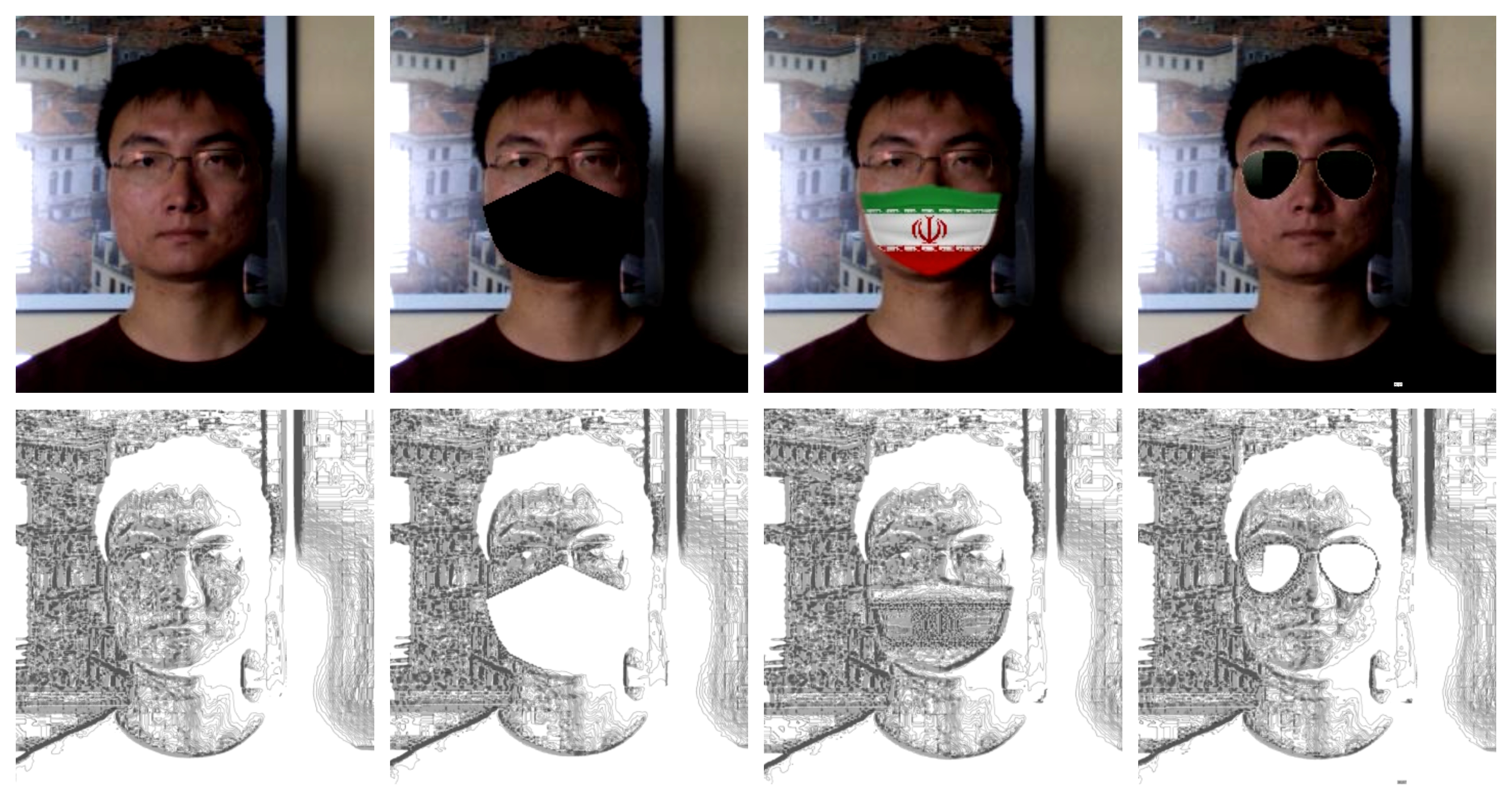}
    \end{center}
    \caption{3 x 3 “uniform” LBP of various occlusions performed on a test image of Replay-Attack~\cite{2} dataset}
    \label{fig:_lbp_}
\end{figure}

In order to understand the rationale of the failure, we investigate one of the classical methods that uses LBP for texture cues and then extends the reasoning to a CNN based method. 
This is demonstrated in Figure~\ref{fig:_lbp_}, where a $3 \times 3$ "uniform" LBP is calculated for an un-occluded and occluded image. It is evident from the figure that a 2D occlusion with a uniform color is creating a texture-less areas on the image whereas, glasses and mask with 3D cues is adding additional texture in either the mouth and nose area or the areas near eyes. Hence, while aggregating the textures in the final feature computation, they differ significantly from a non-occluded image. This analysis can be extended to the CNN architecture as well. As features in a particular part of the face changes, it effects the elements of final feature vector at the end of fully connected layer, hence deteriorating in a similar fashion.  
Adding 3D mask occlusion seems to \textit{fool} algorithms based on handcrafted features, like the LBP+SVM and FD+SVM, into accepting more spoofed images as bonafide as compared to the CNN and Image Quality based model. The Deep-PiXBiS algorithm depicts robustness against false acceptance as we did not see a significant change in the APCER when the model was subjected to mask occlusion attack but, the model suffers from a high BPCER that reduces its overall performance. Even though the CNN based model is showing robustness against false acceptance for 3D mask occlusion attack, the addition of glasses is somehow fooling the algorithm into accepting 20 times more spoofed images as compared to when no occlusion is applied which could be a potential vulnerability in the model. 

\section{Conclusion and future work}
\label{sec:Conclusion}

There are numerous Face Presentation-Attack Detection (PAD) systems currently deployed alongside face recognition algorithms in various smart homes, offices, and industries to prevent unauthorized access by detecting presentation attacks such as the print attack or replay attack. In today's time, the spread of the Corona Virus has made wearing a mask a necessity. Hence, it becomes imperative to re-evaluate the performance of these PAD algorithms when the bonafide individual or the subject in the presentation attack is exposed to mask occlusion. In this paper, we attack four open-source baseline face PAD algorithms with various types of occlusion attacks which include 2D and 3D mask attacks. Additionally, we study the effect of adding glasses to the bonafide and imposter test videos as well. From our analysis, we found that adding such occlusions always deteriorate the overall performance of the baseline algorithms. In most of the rejection to a bonafide person is seen to be increased, which translates to the fact that most of these algorithms will reject any real person if they appear with a mask, the exact scenario of the post pandemic era.
In some cases, especially with 3D masks, we observed that adding occlusions increased the false acceptance rate, which could be a problem for a PAD detection algorithm, especially when critical applications are involved. These behaviors are very much consistent through all spectrum of algorithms including hand crafted features and CNN features. 

Numerous algorithms either have handcrafted features or a CNN architecture at their core and may be susceptible to the addition of such artifacts to the image. Therefore, we have demonstrated that even the best practices in the domain of face PAD could be susceptible to occlusion attacks. Hence, there is an urgent need to revisit the current protocols for face anti-spoofing algorithms. 
An ideal face PAD algorithm should be agnostic to any occlusions such as masks and glasses. Therefore, research efforts can be concentrated on developing new datasets having images of subjects occluded with real occlusions like masks, bandana, mufflers, tattoos, etc., so that PAD algorithms could learn subtle differences between a real-occlusion and a spoofed/synthesized occlusion.


{\small
\bibliographystyle{unsrt}
\bibliography{egbib}
}

\end{document}